\documentclass[sigconf]{acmart}
\usepackage{amsmath}
\usepackage{multirow}
\AtBeginDocument{%
  \providecommand\BibTeX{{%
    \normalfont B\kern-0.5em{\scshape i\kern-0.25em b}\kern-0.8em\TeX}}}

\setcopyright{acmcopyright}
\copyrightyear{2019}
\acmYear{2019}

\begin{document}

\title{SlingDrone: Mixed Reality System for Pointing and Interaction Using a Single Drone}


\author{Evgeny Tsykunov}
\email{Evgeny.Tsykunov@skoltech.ru}
\authornote{Both authors contributed equally to the paper}
\affiliation{%
  \institution{Skoltech, Moscow} 
}

\author{Roman Ibrahimov}
\authornotemark[1]
\affiliation{%
  \institution{Skoltech, Moscow}
}

\author{Derek Vasquez}
\email{Derek.Vasquez@byu.edu}
\affiliation{%
  \institution{BYU, Provo, Utah, USA}
}

\author{Dzmitry Tsetserukou}
\email{D.Tsetserukou@skoltech.ru}
\affiliation{%
  \institution{Skoltech, Moscow}
}





\settopmatter{authorsperrow=4}

\renewcommand{\shortauthors}{Evgeny Tsykunov, et al.}

\citestyle{acmauthoryear}
\begin{abstract}


We propose SlingDrone, a novel Mixed Reality interaction paradigm that utilizes a micro-quadrotor as both pointing controller and interactive robot with a slingshot motion type.
The drone attempts to hover at a given position while the human pulls it in desired direction using a hand grip and a leash. 
Based on the displacement, a virtual trajectory is defined.
To allow for intuitive and simple control, we use virtual reality (VR) technology to trace the path of the drone based on the displacement input.
The user receives force feedback propagated through the leash. 
Force feedback from SlingDrone coupled with visualized trajectory in VR creates an intuitive and user friendly pointing device.
When the drone is released, it follows the trajectory that was shown in VR.
Onboard payload (e.g. magnetic gripper) can perform various scenarios for real interaction with the surroundings, e.g. manipulation or sensing. Unlike HTC Vive controller, SlingDrone does not require handheld devices, thus it can be used as a standalone pointing technology in VR.


\end{abstract}

\ccsdesc[500]{Human-centered computing ~Human computer interaction (HCI)}

\keywords{human-robot interaction, mixed reality, quadrotor, haptics}

\begin{teaserfigure}
  \includegraphics[width=\textwidth]{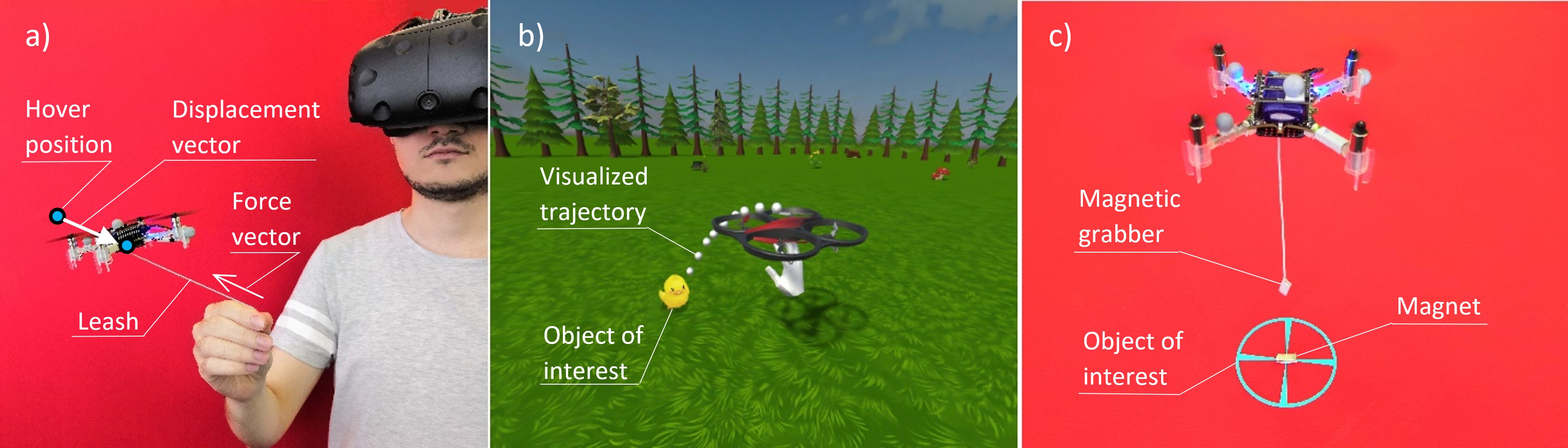}
  \caption{a) The drone points in the slingshot mode: the user pulls the drone using a leash and a hand grip causing a displacement. b) While the drone is in slingshot mode, the trajectory and the object of interest is visualized in virtual scene by simplifying pointing. c) The drone is in projectile mode: it approaches and grabs the object of interest.}
  \label{teaser_fig}
\end{teaserfigure}


\maketitle

\section{Introduction}

Three-dimensional human-computer interaction (HCI) is a rich field which is under active development in the robotics and VR/AR communities.

Pointing during HCI often happens via controllers, such as joysticks or different wearable or handheld devices \cite{Billinghurst_1999, Howard}. For instance, authors in \cite{Speicher_2018} used HTC Vive controllers to select keys on a virtual keyboard. In \cite{yuan2019human}, a human-assisted quadcopter navigation system where the user guides the robot through eye-tracker glasses was proposed. Wearable devices are widely used to control remote \cite{Tsykunov_TOH} or virtual vehicles \cite{Labazanova_2019}. Although different devices proved rich functionality, the user has to hold or wear them during the whole interaction.

One alternative approach is gesture-based interaction which is widely used in VR/AR. \cite{Valentini_2016} showed that Leap Motion is capable of demonstrating suitable accuracy for hand tracking in some operational zones. The drawback is that the finger pointing allows the user to define direction, but not magnitude. Nielsen et al. \cite{Nielsen_2004} stated that performing certain gestures or long lasting gesturing could be stressful. In addition, these approaches in most cases require a camera setup and provide no haptic or tactile feedback. 

Mixed Reality (MR) plays an important role for creating new interaction methods, combining VR and robotics. 
Hoenig et al. \cite{HoenigMixedReality2015} used  MR to simulate bigger quadrotors with small ones, which followed a virtual human. In this way authors tested human-robot interaction algorithms safely and preserved real flight dynamics. Authors in \cite{Freund_1999} used MR for industrial robot teleoperation. A human wore a tracked glove and motions from VR were translated to real manipulators.

In contrast with the discussed works, we introduce a novel MR concept of using a single drone for three-dimensional virtual pointing and interaction. The drone acts as a slingshot and a projectile at the same time. SlingDrone is a selection device which is first used for pointing to an object, then it performs the defined maneuvers to interact with the environment.

The drone has a hand grip hanging from the bottom. By default, the drone hovers at the same position and the human pulls it in some direction using a hand grip, see Fig.\ref{teaser_fig}(a). The physical displacement of the drone (which is three-dimensional) is connected to the virtual pointing and visualized in VR (an example of pointing is shown in Fig.\ref{teaser_fig}(b)). A force feedback system with a leash (Fig.\ref{teaser_fig}(a)) gives the user a better understanding of the interaction. After the pointing phase, the drone is able to perform the defined maneuvers to interact with the surroundings.

In contrast to the GridDrones project \cite{Braley_2018}, where the operator grabs one drone to control the motion of the others agent, the interaction with the drone is smooth by its nature, (i) due to the elastic hardware element (leash) in the interface and (ii) due to the fact that the leash is connected to the center of the drone providing zero angular momentum.

The main novelty of the SlingDrone is that we propose to design and perform maneuvers using one drone, without any additional handheld or wearable control device. For the proof of concept, we tracked the user hand (to be shown in VR to facilitate holder grab) with hand mounted infrared reflective markers, but they can be replaced with Leap Motion attached to the HMD. On the other side, the user also could rely purely on haptic sensation from drone airflow for the hand grab. Since SlingDrone is a flying robot, one of its main advantages is that it can fly to a location that is convenient for a human to perform pointing.





\section{SlingDrone Technology}


A holder allows a human to grab and pull the SlingDrone in any direction in the $X-Y$ plane and downwards (except upwards, which limits the operational zone). 
The drone tries to maintain its desired spatial position \textbf{p\textsubscript{des}} while the operator is pulling the hand grip, which causes a slight change in position.
The state of the SlingDrone, including the current position \textbf{p}, is estimated with onboard sensors and a motion capture system.
The real displacement vector $\textbf{D}$ is defined as $\textbf{p}- \textbf{p\textsubscript{des}}$ and connected to the virtual pointing.

The relation between the real displacement vector and the virtual pointing could have multiple implementations, which incorporates different three-dimensional interaction techniques. For instance, SlingDrone allows to select an object that the user's hand cannot reach. We propose to generate virtual trajectory that starts from the drone position and is defined using the displacement vector $\textbf{D}$. Vector $\textbf{D}$ is tree-dimensional, therefore it is possible to map it to a point in tree-dimensional space.

Force with a vector $\textbf{F}$ distributes through the leash from the drone to the human hand. Based on the slingshot analogy, we assume that force feedback potentially helps to improve interaction and make more accurate pointing.

After the pointing phase, the drone can be transformed to the projectile mode and is able to fly and perform different cases of 3D interaction with the real or virtual environment.

When the drone experiences extreme values in the state parameters (e.g. pitch, roll) during the pointing, the motors shut down. This functionality is introduced as an emergency stop to ensure safety of the operator.

\subsection{Pointing During the Slingshot Mode}

A pointing or the object selection could be defined in multiple ways.
One of the most straightforward solutions is a scaling of displacement vector $\textbf{D}$ with the coefficient $k$ (usually $k>>1$). When the new vector $\textbf{R} = -k\textbf{D}$ defines the point in the mid-air (negative sign also can be omitted, but here it helps to replicate slingshot - we pull in one direction and the projectile fly in the opposite one).
But taking into account that the SlingDrone technology somehow replicates the slingshot operation, we propose to design a ballistic trajectory instead of the straight line. The ballistic trajectory better replicates the projectile flight after the slingshot execution. An intersection of the ballistic trajectory with some surface, line or object could help to define a point in three-dimensional space.

The proposed ballistic trajectory is generated by an air-drag model defined by a system of second order, nonlinear differential equations:

\begin{equation} \label{xydir}  
    \Ddot{x} = \frac{-\rho C_d A_x\dot{x}^2}{2m},
    \Ddot{y} = \frac{-\rho C_d A_y\dot{y}^2}{2m}
\end{equation}

\begin{equation} \label{zdir}  
    \Ddot{z} = \frac{-0.5\rho C_d A_z \dot{z}^2*sin(\dot{z}) - mg}{m}
\end{equation}
where $\rho$ is the air density, $C_d$ is the coefficient of drag, $A$ is the frontal area of the drone in the respective direction, and $m$ is the mass. An air drag model was used to generate an intuitive trajectory of the free ballistic flight.

Equations (\ref{xydir}-\ref{zdir}) are solved using a numerical ODE solver with six initial conditions: the three dimensional vector \textbf{p\textsubscript{des}} as the initial position and the three dimensional displacement vector $\textbf{D}$ as a proxy for initial velocity. $\textbf{D}$ can be scaled by a coefficient $k$ to achieve a desired distance scale for the trajectory. A value of $k=95$ was used. 
The three columns representing position are extracted to form a position vector which defines the trajectory. To command the drone to execute the trajectory, polynomial coefficients were generated as in \cite{richter2016polynomial}. 

The coefficients of equations \eqref{xydir} and \eqref{zdir} have been selected to model a smooth sphere on a ballistic trajectory. The air density $\rho$ is 1.23 kg/m$^3$. The coefficient of drag ($C_d$) is held constant at 0.4. The area is $A_x=A_y=A_z=0.01$m$^2$ and the mass of the virtual body $m$ is 10 kg. Like the trajectory itself, these values can be modified to suit specific use cases.

\subsection{Transition between the Slingshot and the Projectile Modes}
While hovering without interaction with a human, the drone stabilizes itself and experiences small displacements. All displacements under the certain threshold ($\delta_d=20mm$) are not considered as inputs.
During the pointing, the velocity of the drone displacement is small due to the fact that small displacements cause big change in trajectory (which is defined with the $k$ coefficient, described above). In addition, aiming is almost always performed in a slow manner by the user. Therefore, the drone maintains small velocity $v$ which is mostly under the threshold $\delta_v$. Based on that fact, we assume that while the following statements are true
\begin{equation} \label{slingshot_mode_condition}
    \textbf{D}>\delta_d, v<\delta_v
\end{equation}
the drone is in the slingshot mode. We update the trajectory every loop while \eqref{slingshot_mode_condition} is true. When the user releases the holder, the drone starts to accelerate towards the default hover position and the velocity $v$ becomes bigger than $\delta_v$ (the trajectory is not updated any more). When the drone approaches the hover position $\textbf{D}<\delta_d$, it becomes a projectile and starts to follow the last valid trajectory towards the defined point in space to interact.

\subsection{Human-SlingDrone Interaction Strategy}
First, the SlingDrone approaches the human and starts to hover, entering the slingshot mode (Fig.\ref{teaser_fig}(a)). The user estimates the position of the drone utilizing visual feedback from VR coupled with a haptic sensation form the airflow below the quadrotor, which helps to catch the hand grab. After the human grabs the holder, he or she starts to pull it. When the magnitude of the current displacement vector |$\textbf{D}$| exceeds the $\delta_d$ threshold, the trajectory is being visualized in VR. Now, the human is able to point the SlingDrone by pulling the holder.
When the user is satisfied with the pointing or the selection, he can release the holder. The drone enters the projectile mode (Fig.\ref{teaser_fig}(c)) and performs the maneuvers and interaction with the surroundings.

\subsection{Implementation}

The SlingDrone itself, together with the user's hand, and the object of interest are real objects and tracked by a motion capture system with visualization in virtual scene.

\subsubsection{Aerial Platform}
We used a Crazyflie 2.0 quadrotor to perform the verification of the flight tests.
To get the high-quality tracking of the quadrotor during the experiments, we used Vicon motion capture system with 12 cameras (Vantage V5) covering a $5m \times  5 m \times  5 m$ space. We used the Robot Operating System (ROS) Kinetic framework to run the developed software and ROS stack \cite{Preiss_2017, HoenigMixedReality2015} for Crazyflie 2.0. The position and attitude update rate was 100 Hz for all drones. Before the verification of the proposed approach, we ensured that we were able to perform a stable and smooth flight, following the desired trajectory. In order to do so, all PID coefficients for position controller were set to default values for Crazyflie 2.0, according to \cite{Preiss_2017} (for x,y-axis $k_p$=0.4, $k_d$=0.2, $k_i$=0.05; for z-axis $k_p$=1.25, $k_d$=0.4, $k_i$=0.05). Finally, to be able to reach stable displacement of the SlingDrone during interaction, we set all positional integral terms of the PID controller to zero.

An elastic wire (100 mm in length) is connected to the bottom of the SlingDrone, as shown in Fig.\ref{teaser_fig}(a), with a holder at the end. 

\subsubsection{VR Development}
A Virtual Reality application in Unity3D was created to provide the user with a pleasant immersion. Quadcopter, human hand, and remote object models were  simulated  in  VR. The  size  of  each  virtual  object changes like in real life depending on the distance between the operator and the objects. Vicon Motion Capture cameras were used to transfer positioning and rotation tracking data of each object to the Unity 3D engine. The operator wears HTC Vive VR headset to experience Virtual Reality application.
 
The virtual trajectory is updated 30 times per second and the human can get visual feedback in VR about the potential flight path (to avoid obstacles) and about the destination point.

\begin{figure*}[t]
\centering
\includegraphics[width=\textwidth]{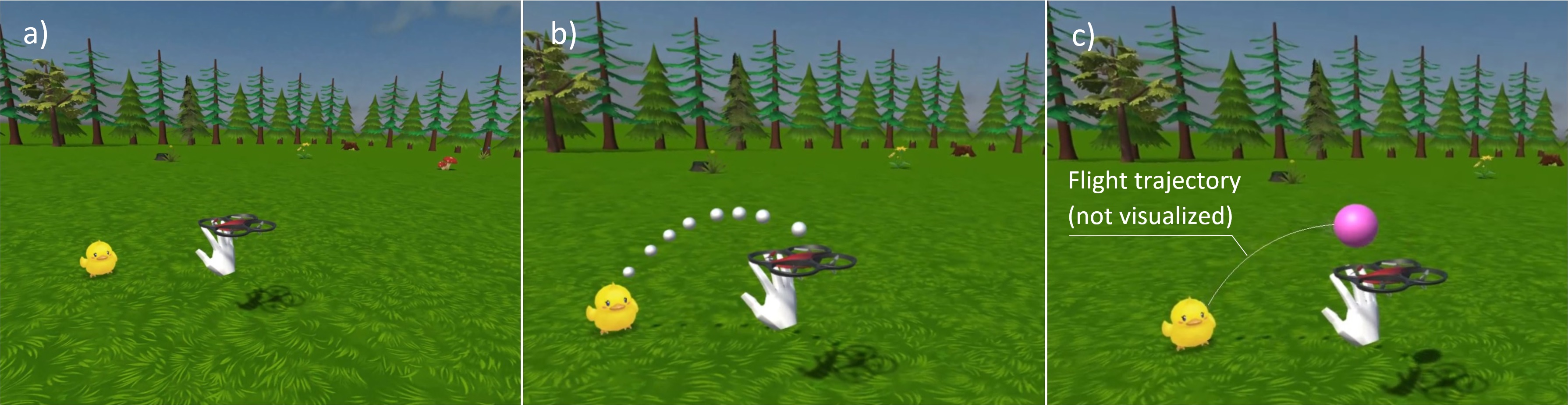}\label{true}
\caption{Workflow of the user study experiment. (a) No interaction between the human and the drone, (b) User points towards the object of interest, (c) Pink ball represents the projectile.}
\label{fig2}
\end{figure*}

\subsection{Application: Grabbing a Remote Object}
The goal of the proposed application is to grab a remote object using a magnetic gripper placed on the Crazyflie 2.0 drone. As a gripper for the user, we used a magnet which is also used to attract the remote object of interest.

During the pointing, when the end point of the generated trajectory hits the remote object that is intended to be picked, the user releases the holder. The drone enters the projectile mode (Fig.\ref{teaser_fig}(c)) and starts to follow the defined trajectory representing the slowed flight of passive object. When the quadrotor approaches the end of trajectory, to increase the chances of grabbing, it performs a search of object of interest by moving in a square with 0.15 meter side. After that, the drone flies back to the user to perform a delivery.
After the delivery the user is able to experience a tangible sensation of the object of interest.

\section{User Study}
Seven right-handed users (22 to 28 years old, six male and one female) took part in the experiments. 
The protocol of the experiment was approved by a Skolkovo Institute of Science and Technology review board and all participants gave informed consent.

\subsection{Experimental Methods}
As described above, the user's hand, the drone, and the object of interest are visualized in VR. The drone hovered near the human and maintained 1.5 meters in height as shown in Fig. \ref{fig2}(a). For the experiment we decided to focus on the pointing functionality of the SlingDrone purely in VR. Therefore, we kept the slingshot mode of the drone for the whole experiment. After the user performed aiming (with visualized trajectory as shown in Fig. \ref{fig2}(b)) and released the hand grip, the drone avoided the transition to the projectile mode. Instead, after the release, the drone just moved to the default hover position. Then, the virtual ball appeared and became a projectile; then, it flied along the desired trajectory in virtual scene (Fig. \ref{fig2}(c)).
All subjects were asked to evaluate the pointing capabilities of SlingDrone technology for 5 minutes. Pointing is just a part of the idea behind SlingDrone (another part is the physical flight for interaction), which means that the experimental results cannot be extrapolated to the whole technology.

\subsection{Experimental Results and Discussion}

All participants positively responded to the device convenience.
After the experiment, we also asked the subjects to answer a questionnaire of 8 questions using bipolar Likert-type seven-point scales. The results are presented in Table \ref{table}.

All participants reported that it is was easy to learn of how to use the SlingDrone (in fact it took 10-30 second to understand the workflow), which is supported by question 1 in Table \ref{table}. 
According to the users, it was slightly hard to grab the holder, as far as it is not visualised in VR. For the future work, the holder will be visualised in VR.
For the trajectory representation, we used small white balls thrown from the position of the drone. Therefore, when the users change the direction, the trajectory did not change instantly - it takes some time for the balls to perform a flight. That fact was not positively supported by the users. For the future, we plan to replace the balls with a trajectory visualised with dashed line that can be changed instantly.
Surprisingly, in spite of the small size of the used quadrotor, most of the users reported that they felt some force feedback from the drone, which provides additional information about the magnitude of the input.

In question 5 users were satisfied with the actual destination of the virtual projectile. Along with that, pulling a drone looks similar to the slingshot operation, which could have an acceptable accuracy in certain cases. Therefore, we propose a hypothesis that the SlingDrone potentially could have a high accuracy. During the experiment we did not measure the accuracy of the pointing, but it will be done in the future work.

\begin{table}[]
\caption{Evaluation of the participant's experience. Volunteers evaluated these statements, presented in random order, using a 7-point Likert scale (1 = completely disagree, 7 = completely agree). Means and standard deviations are presented.}
\label{table}
\begin{tabular}{|l|l|c|c|}
\hline
 & \multicolumn{1}{c|}{\textbf{Questions}} & \multicolumn{1}{l|}{\textbf{Mean}} & \multicolumn{1}{l|}{\textbf{SD}} \\ \hline
1 & It was easy to learn how to use SlingDrone. & 6.4 & 0.49 \\ \hline
2 & \begin{tabular}[c]{@{}l@{}}It was easy to grab and keep the holder of \\ the drone.\end{tabular} & 5.4 & 1.02 \\ \hline
3 & \begin{tabular}[c]{@{}l@{}}The visualized trajectory in VR clearly \\ represented the flight path.\end{tabular} & 5.6 & 0.8 \\ \hline
4 & \begin{tabular}[c]{@{}l@{}}I felt the force response from the drone \\ when pulling the rope.\end{tabular} & 5.8 & 1.47 \\ \hline
5 & \begin{tabular}[c]{@{}l@{}}I was satisfied with the actual destination \\ of the virtual projectile.\end{tabular} & 5.2 & 0.74 \\ \hline
6 & I was tired at the end of the experiment. & 1.4 & 0.48 \\ \hline
7 & I felt comfortable when I used SlingDrone. & 6.0 & 0.89 \\ \hline
8 & I felt safe when I used SlingDrone. & 7.0 & 0 \\ \hline
\end{tabular}
\end{table}

\section{Conclusion and Future Work}\label{conclusion}
In this paper, we developed SlingDrone - a novel interaction technology that combines the advantages of real and virtual environments. One of the core features is that SlingDrone provides a powerful and easy to use 3D pointing technology.
Therefore, SlingDrone can also be used for pointing of other robots. The intersection between the virtual trajectory and a floor can be used as a desired position for a ground based robot such as car.

As a perspective development, we are planning to introduce the SlingDrone placement mode, when the user moves the SlingDrone in space to the most suitable location by pulling the rope. This can be done as a preliminary setup before the actual pointing process.

To provide deeper immersion into the pointing process, instead of the magnetic hand grab, the vibromotor can be used. The vibromotor is connected to the drone via coiled elastic wire and it is powered/controlled from onboard electronics. Different tactile patterns can deliver additional information about the potential interaction with the surroundings. But for the selection applications a press button instead of the vibromotor can be more suitable.


For the future work, we plan to conduct a user study to evaluate the participants' experience of using SlingDrone and to estimate the accuracy and precision of pointing and flight. 
The SlingDrone platform can also be implemented in an Augmented Reality (AR). This would allow for more direct information about the trajectory to the goal and provide additional safety. 

\bibliographystyle{ACM-Reference-Format}
\bibliography{sample-base}

\end{document}